%% file: main.tex
\documentclass[letterpaper]{article} 
\usepackage{aaai2026}  
\usepackage{times}  
\usepackage{helvet}  
\usepackage{courier}  
\usepackage[hyphens]{url}  
\usepackage{graphicx} 
\usepackage{natbib}  
\usepackage{caption} 
\usepackage{algorithm}

\usepackage{newfloat}
\usepackage{listings}
\DeclareCaptionStyle{ruled}{labelfont=normalfont,labelsep=colon,strut=off} 
\floatstyle{ruled}
\newfloat{listing}{tb}{lst}{}
\floatname{listing}{Listing}
\usepackage{amsmath}
\usepackage{amssymb}
\usepackage{amsfonts}
\usepackage{booktabs}
\usepackage{amsfonts}
\usepackage{bm} 
\usepackage{algpseudocode} 
\usepackage{svg}
\usepackage{amsmath}
\usepackage{xspace} 
\usepackage{pifont}
\usepackage{subcaption}
\usepackage{adjustbox}
\usepackage{colortbl}
\usepackage{algorithm}
\usepackage{algpseudocode}
\usepackage{multirow}
\usepackage{xcolor}
\usepackage{units}
\usepackage{amsfonts,amssymb}
\usepackage{caption}
\usepackage{graphicx}
\usepackage{multirow}
\usepackage{amsmath,amssymb,amsthm}
\usepackage{graphicx}
\usepackage{multirow}
\usepackage{amsmath,amssymb,amsthm}

\usepackage{ragged2e}

\usepackage{url}            
\usepackage{colortbl}

\usepackage{array}
\usepackage{tabularx}
\usepackage{makecell}

\newcolumntype{C}[1]{>{\Centering\arraybackslash}p{#1}}

\usepackage{booktabs}
\usepackage{xcolor}
\definecolor{mycolor}{rgb}{0.5,0.1,0.8} 
\usepackage{xspace}
\usepackage{xcolor}
\usepackage{tikz}
\usepackage{titletoc}
\usepackage{svg}
\usepackage{amsmath}
\usepackage{soul}

\newcommand{\algname}{D²Pruner\xspace}

\newcommand{\fg}[1]{\mathbf{\mathcolor{ForestGreen}{#1}}}

\nocopyright
\title{D²Pruner: Debiased Importance and Structural Diversity \\for MLLM Token Pruning}
\author{
    Evelyn Zhang\textsuperscript{\rm 1,\rm 2}\thanks{This work was done during Evelyn Zhang's internship at Tencent YouTu Lab.},
    Fufu Yu\textsuperscript{\rm 1},
    Aoqi Wu\textsuperscript{\rm 3},
    Zichen Wen\textsuperscript{\rm 2},
    Ke Yan\textsuperscript{\rm 1},
    Shouhong Ding\textsuperscript{\rm 1}, 
    Biqing Qi\textsuperscript{\rm 4},
    Linfeng Zhang\textsuperscript{\rm 2}\thanks{corresponding author}
}
\affiliations{
    \textsuperscript{\rm 1}Tencent YouTu Lab\\
    \textsuperscript{\rm 2}School of Artificial Intelligence, Shanghai Jiao Tong University\\
    \textsuperscript{\rm 3}School of Computer Science and Technology, Tongji University\\
    \textsuperscript{\rm 4}Shanghai AI Laboratory\\

    zhanglinfeng@sjtu.edu.cn
}

\usepackage{bibentry}

\begin{document}

\maketitle
\begin{abstract}
Processing long visual token sequences poses a significant computational burden on Multimodal Large Language Models (MLLMs). While token pruning offers a path to acceleration, we find that current methods, while adequate for general understanding, catastrophically fail on fine-grained localization tasks. We attribute this failure to the inherent flaws of the two prevailing strategies: importance-based methods suffer from a strong positional bias, an inherent model artifact that distracts from semantic content, while diversity-based methods exhibit structural blindness, disregarding the user's prompt and spatial redundancy. To address this, we introduce D²Pruner, a framework that rectifies these issues by uniquely combining debiased importance with a structural pruning mechanism. Our method first secures a core set of the most critical tokens as pivots based on a debiased attention score. It then performs a Maximal Independent Set (MIS) selection on the remaining tokens, which are modeled on a hybrid graph where edges signify spatial proximity and semantic similarity. This process iteratively preserves the most important and available token while removing its neighbors, ensuring that the supplementary tokens are chosen to maximize importance and diversity. Extensive experiments demonstrate that D²Pruner has exceptional efficiency and fidelity. Applied to LLaVA-1.5-7B for general understanding tasks, it reduces FLOPs by 74.2\% while retaining 99.2\% of its original performance. Furthermore, in challenging localization benchmarks with InternVL-2.5-8B, it maintains 85.7\% performance at a 90\% token reduction rate, marking a significant advancement with up to 63. 53\% improvement over existing methods. 

\end{abstract}
\begin{links}
    \link{Code}{https://github.com/EvelynZhang-epiclab/D2Pruner}
\end{links}

\input{section/1_introduction}

\input{image/overview}
\input{section/2_related_work}

\input{section/3_method}
\input{section/4_experiment}

\input{section/5_conclusion}
\bibliography{aaai2026}
\newpage
\appendix

\end{document}

%% file: section/1_introduction.tex
\section{Introduction}
The remarkable capabilities of Multimodal Large Language Models (MLLMs) in visual understanding and localization are fundamentally constrained by their computational demands. The quadratic complexity of self-attention and the massive memory footprint of the Key-Value (KV) cache create a severe bottleneck, particularly when processing high-resolution images or long videos. This bottleneck critically hinders their efficient inference and real-world deployment, making inference acceleration an urgent priority.
\input{image/modelbias}
\input{image/motivation}

To mitigate this, token pruning has emerged as a promising strategy for reducing the input sequence length. While existing methods show moderate success on general understanding and reasoning tasks, they catastrophically fail on fine-grained tasks that demand precise spatial awareness and localization. Our investigation pinpoints two critical, previously overlooked failure modes in these approaches:

\textbf{Position Bias in Importance-based Methods.} These methods~\cite{fastv} estimate the importance of visual tokens by leveraging the attention scores assigned to them by the last token. However, such scores are inherently vulnerable to position bias, which can distort the true importance distribution. Prior studies~\cite{wen2025token,fastervlm} have reported that LLaVA consistently exhibits a content-agnostic emphasis on the bottom region of the image. Building on this observation, we conduct a broader analysis across several representative MLLMs and reveal that position bias is not limited to LLaVA but manifests in diverse forms. Specifically, we average the attention maps of over 1,000 randomly sampled images from the COCO dataset~\cite{coco} to uncover these patterns. As highlighted in Fig.\ref{fig:modelbias}, our findings not only corroborate LLaVA’s bottom-focused tendency but also expose distinct biases in other models: InternVL\cite{internvl2.5} and QwenVL~\cite{qwenvl2.5}, for instance, exhibit a pronounced preference for tokens near the image periphery. We posit these biases stem from the models' positional embeddings. LLaVA's \textit{absolute embeddings} foster a location-based shortcut (bottom-focus), while the \textit{relative embeddings} (e.g., RoPE~\cite{rope}) in InternVL and QwenVL encourage a structural one (periphery-focus). This persistent position bias fundamentally undermines the reliability of raw attention as a proxy for semantic importance.

\textbf{Structural Blindness in Diversity-based Methods. }Previous methods~\cite{dart,divprune} prune tokens by assessing diversity purely in the feature space, typically via cosine similarity. This strategy operates on a flawed premise: that visual information can be processed as a one-dimensional sequence, after flattening the original 2D grid. This action strips away all topological information, forcing the model to treat tokens as an unstructured set and rendering it blind to their crucial spatial relationships. This motivates our approach to adopt a representation that explicitly preserves the 2D structure of images by modeling tokens and their semantic–spatial relationships as a graph.

To address these challenges, we introduce \textbf{D²Pruner}, a pruning framework built on the principles of \textbf{D}ebiased Importance (DI) and Structural \textbf{D}iversity (SD). It first secures a set of the most critical tokens as pivots based on the debiased attention score. It then performs a greedy Maximal Independent Set (MIS) selection on the remaining tokens, which are modeled on a hybrid graph where edges signify both spatial proximity and semantic similarity. This process iteratively preserves the most important available token while pruning its neighbors, ensuring the supplementary tokens are chosen to maximize importance and structural diversity. 

Leveraging its dual-stage design, D²Pruner not only excels on general understanding tasks but demonstrates unparalleled strength in fine-grained localization, as illustrated in Fig.\ref{fig:motivation}. On challenging localization benchmarks~\cite{refcoco,refcocog}, our method maintains 85.7\% performance on InternVL-2.5-8B~\cite{internvl2.5} even at a 90\% pruning rate—surpassing prior diversity-based and importance-based methods by a massive 63.5 and 51.2 points, respectively. This is complemented by strong general performance, retaining 95.5\% of LLaVA-Next-7B's capabilities. Critically, these results are achieved with remarkable efficiency: D²Pruner delivers a 5× prefill acceleration and a 7.6× reduction in KV cache size, while the pruning process itself adds a negligible overhead of just 0.8\% to the forward pass. In summary, our contributions are as follows.
\begin{itemize}
    \item We analyze the pervasive position bias in various MLLMs and reveal distinct bias patterns across models. To address this, we propose a debiasing method that effectively mitigates bias in token importance estimation.
    \item We present D²Pruner, a token pruning framework that jointly considers both spatial and semantic redundancy by modeling token relationships on a hybrid graph, enabling more informed and effective pruning decisions.
    \item Extensive experiments demonstrate that D²Pruner achieves superior performance, especially on fine-grained localization tasks, while delivering significant acceleration and memory savings.
\end{itemize}

%% file: image/modelbias.tex
\begin{figure}[t]
\centering
\small
\includegraphics[width=0.48\textwidth]{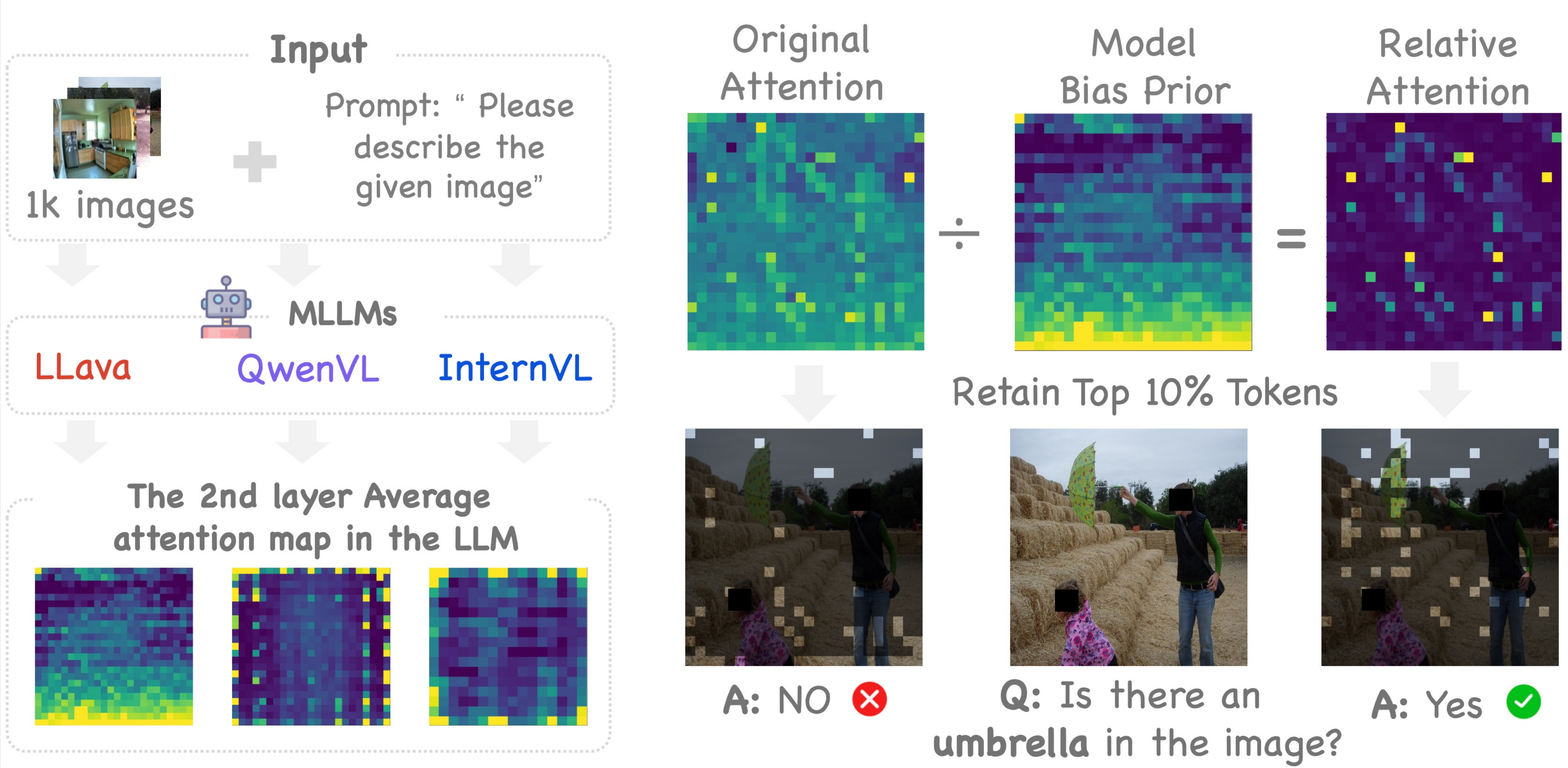}
\vspace{-10pt}
\caption{\textbf{Illustration of positional bias in MLLMs and our debiasing solution.} \textit{Left}: Averaging attention maps over 1k images reveals a strong model bias prior (e.g., towards the bottom for LLaVA).   \textit{Right}: This causes the naive attention-based method (e.g., FastV) to fail by pruning salient objects like the umbrella. Our relative attention, which normalizes by this model bias prior, rectifies the pruning decision and answer correctly.}
\vspace{-10pt}
\label{fig:modelbias}
\end{figure}

%% file: image/motivation.tex
\begin{figure*}[t]
\centering
\small
\includegraphics[width=1\textwidth]{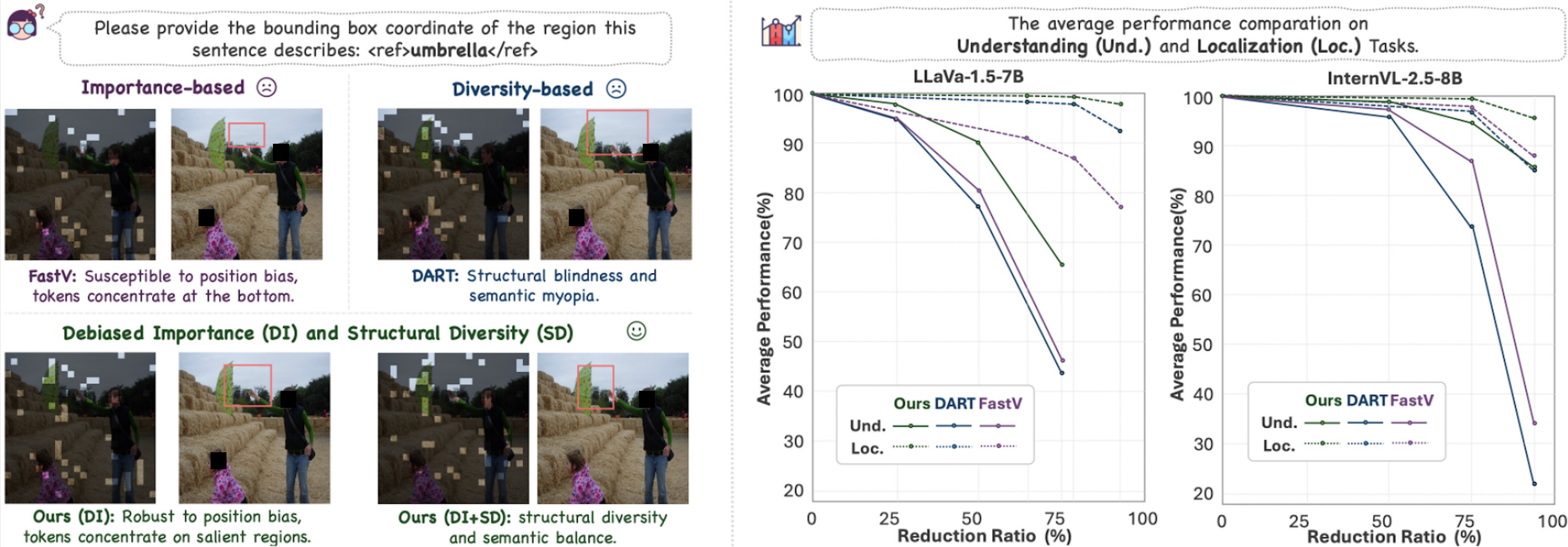}
\vspace{-8pt}
\caption{\textbf{Qualitative and Quantitative Analysis of D²Pruner against Prior Methods.} \textit{Left:} Visual comparison on localization tasks. \textcolor[HTML]{512B56}{Importance-based method, FastV}, is misled by positional bias, concentrating tokens on the image bottom, while \textcolor[HTML]{213D61}{Diversity-based method, DART}, suffers from structural blindness and semantic myopia. In contrast, our \textcolor[HTML]{224C23}{Debiased Importance (DI)} component successfully overcomes positional bias. Our full model, \textcolor[HTML]{224C23}{D²Pruner (DI+SD)}, jointly addresses both semantic and spatial redundancy, resulting in a more balanced and informative token selection that enhances localization performance. \textit{Right:} Quantitative evaluation of average performance on general understanding and fine-grained localization tasks. D²Pruner achieves the best performance across all reduction ratios, with notable improvements on localization tasks.}
\vspace{-10pt}
\label{fig:motivation}
\end{figure*}


%% file: image/overview.tex
\begin{figure*}[t]
\centering
\small
\includegraphics[width=1\textwidth]{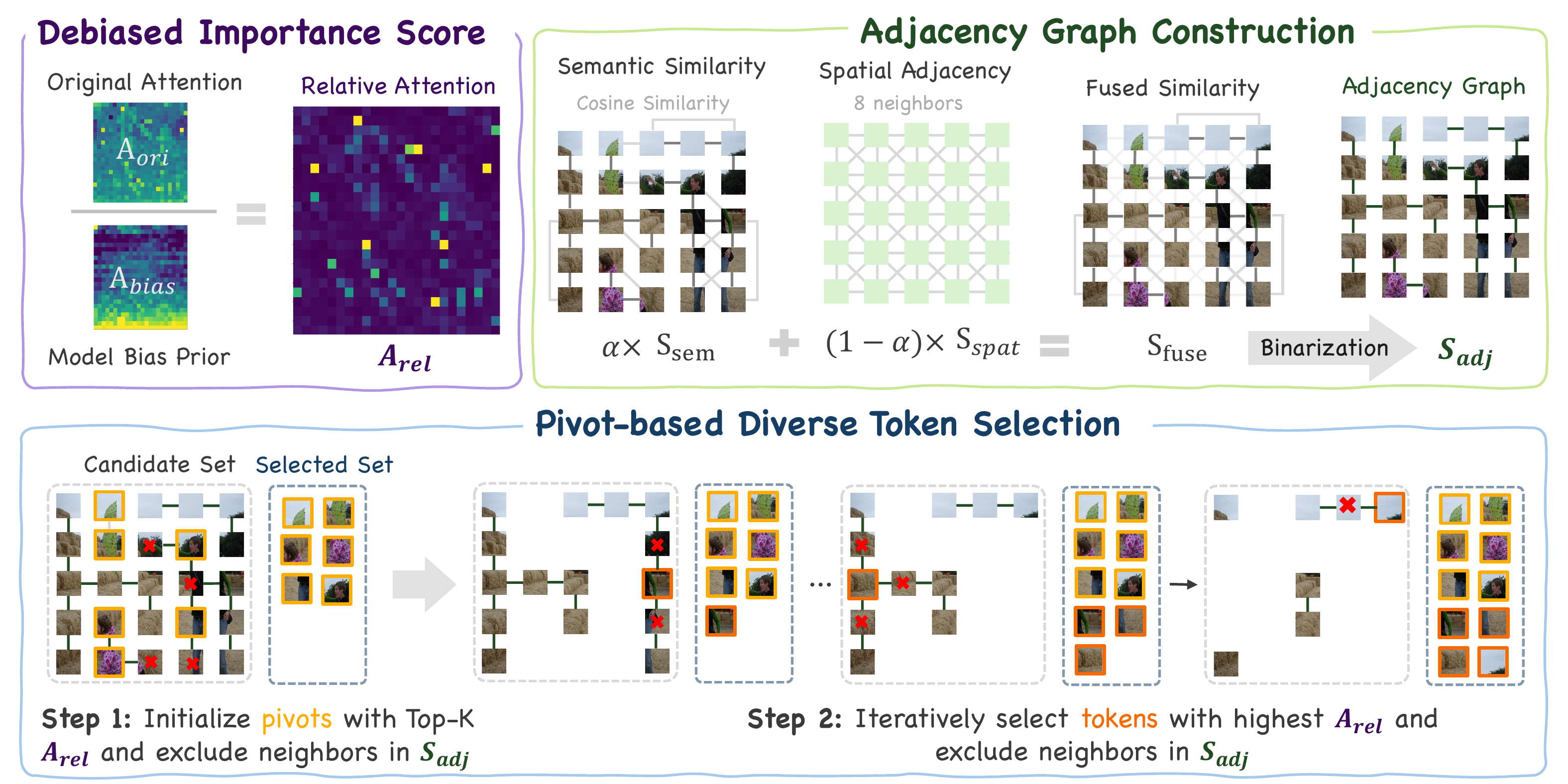}
\vspace{-8pt}

\caption{\textbf{Overall pipeline for our proposed D²Pruner. } \textit{\textcolor[HTML]{512B56}{(a) Debiased Importance Score:}} The original attention map $A_{ori}$ is adjusted by subtracting the model bias prior $A_{bias}$, resulting in the relative attention map $A_{rel}$, \textit{\textcolor[HTML]{224C23}{(b) Adjacency Graph Construction:}} The semantic similarity and spatial adjacency are fused to create the adjacency graph $S_{adj}$. \textit{\textcolor[HTML]{213D61}{(c) Pivot-based Diverse Token Selection:}} In Step 1, pivots are initialized using the top-K values from $A_{rel}$, excluding neighbors as defined by $S_{adj}$; In Step 2, tokens are iteratively selected based on the highest $A_{rel}$ values, with neighbors excluded according to the adjacency graph. }
\vspace{-10pt}
\label{fig:overall}
\end{figure*}

%% file: section/2_related_work.tex
\section{Related Work}
\subsection{Multimodal Large Language Models}
\label{sec:related_work_mllm}

The dominant architecture for Multimodal Large Language Models (MLLMs), pioneered by LLaVA~\citep{llava}, connects a pre-trained vision encoder to a Large Language Model (LLM) via a lightweight projection module. This paradigm has spurred rapid advancements, with successors like LLaVA-NeXT~\citep{llavanext}, Qwen-VL~\citep{qwenvl}, and InternVL~\citep{internvl} pushing the state-of-the-art. These models achieve superior performance by leveraging more powerful foundation models, scaling up training data, and enhancing their capability to process high-resolution images.
However, this pursuit of higher performance, especially through high-resolution inputs, introduces a significant computational burden. The number of visual tokens grows quadratically with image resolution, creating a major bottleneck in the LLM's self-attention layers. Our work is directly motivated by the need to alleviate this bottleneck through efficient visual token compression, making these powerful models more accessible and practical.

\subsection{Visual Token Compression}
\label{sec:related_work_compression}

To mitigate the high computational cost of MLLMs, visual token pruning aims to reduce the number of visual tokens fed to the language model. Existing methods can be broadly classified into two main paradigms: importance-based and diversity-based pruning.

\textbf{(1) Importance-Based Pruning.}
Early and prominent methods operate on the principle of importance-based pruning, where tokens with the highest scores are retained. Some methods~\cite{fastv,sparsevlm,pdrop} leverage the attention scores assigned to visual tokens by the last token within the language model to evaluate the importance of each visual token. However, such methods suffer from positional bias, often favoring tokens at the bottom of image regardless of content~\cite{wen2025token,fastervlm}. Other methods~\cite{llava-prumerge,visionzip,fastervlm} rely on the attention scores assigned to visual tokens by class token in the vision encoder, suffering from instruction-agnostic token selection.

\textbf{(2) Diversity-Based Pruning.}
Recent studies have moved beyond importance-based token pruning, highlighting the benefits of reducing redundancy through diversity-based strategies. Approaches such as DivPrune~\cite{divprune} formulate token selection as a diversity maximization problem, ensuring that the retained tokens are highly representative of the original set. Similarly, DART~\cite{dart} demonstrates that importance scores may not be reliable indicators for pruning and instead proposes a duplication-aware method, which selects tokens with minimal overlap to a set of pivots. But, these methods only consider semantic redundancy (e.g., cosine similarity) and often exhibit \textit{structural blindness} by ignoring spatial relationships among tokens. They also fail to incorporate user instructions, limiting their adaptability.

%% file: section/3_method.tex
\section{Methodology}
\label{sec:methodology}
In this section, we present \textbf{D²Pruner}, a training-free framework designed to accelerate the inference of MLLMs. D²Prune is built on two key principles: debiased importance and structural diversity, as illustrated in Fig.~\ref{fig:overall}.

\subsection{Preliminary and Problem Formulation}
\label{sec:problem_formulation}

An MLLM typically comprises a vision encoder $\mathcal{E}_v$ and a Large Language Model (LLM) $\mathcal{L}$. Given an image $I$, $\mathcal{E}_v$ produces a sequence of $N$ visual tokens, $V = \{v_1, \dots, v_N\}$, where each $v_i \in \mathbb{R}^d$. These are prepended to $M$ text tokens $T$ to form the input sequence $X = [V; T]$ for the LLM. The self-attention mechanism's quadratic complexity, $O((N+M)^2)$, makes the large number of visual tokens a primary computational bottleneck.

Token pruning aims to mitigate this cost by selecting an informative subset of visual tokens. We perform this pruning dynamically within the LLM's architecture. Let $H^{(k)}$ be the sequence of hidden states at the output of the $k$-th transformer layer. We apply our pruner at a pre-determined depth, after layer $K-1$. The pruner operates on the visual component of the hidden states, $H_V^{(K-1)}$, to produce a condensed set $H'_{V}{}^{(K-1)}$ of size $n = r \cdot N$ for a given pruning ratio $r$. This pruned set is then concatenated with the untouched text hidden states, $H_T^{(K-1)}$, to form the input $H'^{(K-1)} = [H'_{V}{}^{(K-1)}; H_T^{(K-1)}]$ for all subsequent layers $k \ge K$.
We aim to select a subset of hidden states $H_V^{(K-1)} = \{h_1, \dots, h_N\}$ that is maximally informative and representative. We define our objective as finding the selection mask $\mathbf{s}^*$ that maximizes a composite score function $\mathcal{F}$:
\begin{equation}
\label{eq:proxy_objective}
\mathbf{s}^* = \arg\max_{\mathbf{s}} \mathcal{F}(H_V^{(K-1)}, \mathbf{s}) \quad \text{subject to} \quad \sum_{i=1}^{N} s_i = n
\end{equation}
The score function $\mathcal{F}$ is designed to explicitly reward both individual token importance and overall subset diversity:
\begin{equation}
\label{eq:score_function}
\mathcal{F}(H_V^{(K-1)}, \mathbf{s}) = \underbrace{\sum_{i=1}^{N} s_i \cdot \mathcal{I}(h_i)}_{\text{Token Importance}} + \lambda \cdot \underbrace{\text{Div}(\{h_i | s_i=1\})}_{\text{Subset Diversity}}
\end{equation}
where $\mathcal{I}(h_i)$ is the debiased Importance score of token $h_i$, quantifying its intrinsic value. $\text{Div}(\cdot)$ is the structural diversity function, measuring the structural comprehensiveness and low redundancy of the selected subset. $\lambda$ is a hyperparameter balancing the trade-off between importance and diversity.
The following sections will detail our methods for computing $\mathcal{I}(h_i)$ and for designing an efficient selection algorithm that jointly optimizes this objective.

\input{table/llava_v1.5}

\input{table/ud}
\input{table/vg}
\subsection{Debaised Importance Score}
\label{sec:bias_free_importance}

The first term in our objective function (Eq. \ref{eq:score_function}), Total Importance, relies on computing a robust and reliable importance score $\mathcal{I}(h_i)$ for each visual token's hidden state $h_i$. A naive reliance on raw attention scores is prone to architectural artifacts, such as positional biases, where tokens at certain positions (e.g., the end of a sequence) receive artificially high attention regardless of their semantic content. 
We leverage a novel \textit{relative attention score}, which is specifically designed to disentangle true semantic salience from systemic positional bias. We define this score as follows:

First, for a given image-text input, we extract the raw attention weights directed from the final text token to each of the $N$ visual tokens at layer $K-1$. Let this vector of content-specific attention scores be $\mathcal{A}_{ori} \in \mathbb{R}^N$, where $\mathcal{A}_{ori}(i)$ reflects the model's focus on visual token $h_i$ for the current task.
Second, to isolate the inherent positional bias, we conduct a one-time calibration to compute a positional bias prior, denoted as $\mathcal{A}_{bias} \in \mathbb{R}^N$. This prior is obtained by feeding the model 1,000 randomly selected images from the COCO dataset, paired with a generic, non-specific prompt (e.g., "Please describe the provided image."). The final attention map is then averaged over all 1,000 images.
The resulting attention scores, $\mathcal{A}_{bias}(i)$, capture the model's default attention distribution across the visual token positions when no meaningful visual information is present.

The final relative attention score, $\mathcal{A}_{rel}$, is then calculated by normalizing the content-specific attention by this positional bias prior:
\begin{equation}
\label{eq:relative_attention}
\mathcal{A}_{rel} = \frac{\mathcal{A}_{ori}}{\mathcal{A}_{bias} + \epsilon}
\end{equation}
where $\epsilon$ is a small constant (e.g., $10^{-7}$) to ensure numerical stability. Our method also supports dynamic input resolutions. Specifically, when the number of visual tokens varies (e.g., due to different image resolutions or patch sizes), we resize the positional bias prior $\mathcal{A}_{bias}$ to match the size of $\mathcal{A}_{ori}$ using interpolation. This ensures $\mathcal{A}_{rel} $ remains consistent and applicable across varying input resolutions.

\subsection{Adjacency Graph Construction}

Let the set of visual token features be denoted by $H_V = \{h_1, h_2, \dots, h_N\}$, where $h_i \in \mathbb{R}^D$ is the $D$-dimensional feature vector for the $i$-th token. These $N$ tokens correspond to a spatial grid of patches of size $h \times w$, such that $N = h \times w$. Our goal is to compute an adjacency matrix $\mathcal{S}\in \{0, 1\}^{N \times N}$ that defines the connectivity of the graph.

To capture the content-based relationships between tokens, we compute the pairwise cosine similarity. The semantic similarity $S_{\text{sem}}(i, j)$ between two token features $h_i$ and $h_j$ is calculated as:
\begin{equation}
    S_{\text{sem}}(i, j) = \frac{h_i \cdot h_j}{\|h_i\| \|h_j\|}
    \label{eq:cosine_sim}
\end{equation}
This computation is performed for all pairs of tokens, yielding a dense semantic similarity matrix $\mathbf{S}_{\text{sem}} \in \mathbb{R}^{N \times N}$. 
Then, we apply min-max normalization to the semantic similarity matrix for normalization:
\begin{equation}
    \hat{\mathbf{S}}_{\text{sem}} = \frac{\mathbf{S}_{\text{sem}} - \min(\mathbf{S}_{\text{sem}})}{\max(\mathbf{S}_{\text{sem}}) - \min(\mathbf{S}_{\text{sem}})}
    \label{eq:normalization}
\end{equation}

To re-introduce the spatial structure, we define a spatial proximity matrix $\mathbf{S}_{\text{spat}}$ based on adjacency in the original image grid. An edge is considered to exist between two tokens if their corresponding patches are immediate neighbors. We employ an 8-connectivity rule. Formally, the spatial proximity matrix $\mathbf{S}_{\text{spat}} \in \{0, 1\}^{N \times N}$ is defined as:
\begin{equation}
    S_{\text{spat}}(i, j) = 
    \begin{cases} 
        1 & \text{if } j \text{ is in the 8-neighborhood of } i \\
        0 & \text{otherwise}
    \end{cases}
    \label{eq:spatial_sim}
\end{equation}
This results in a sparse, binary matrix that exclusively encodes the local spatial topology.

Subsequently, the two normalized matrices are fused via a weighted sum to produce a final similarity matrix $\mathbf{S}_{\text{fused}}$:
\begin{equation}
    \mathbf{S}_{\text{fused}} = \alpha \cdot \hat{\mathbf{S}}_{\text{sem}} + (1 - \alpha) \cdot \mathbf{S}_{\text{spat}}
    \label{eq:fusion}
\end{equation}
where $\alpha \in [0, 1]$ is a hyperparameter that balances the contribution of semantic content versus spatial structure.

Finally, the fused similarity matrix $\mathbf{S}_{\text{fused}}$ is converted into a binary adjacency matrix $\mathcal{S}$ by applying a similarity threshold, $\theta_{\text{sim}}$. An edge is created between two nodes if their fused similarity exceeds this threshold. The elements of the final adjacency matrix $\mathbf{S}$ are defined as:
\begin{equation}
    \mathcal{S}(i, j) = 
    \begin{cases} 
        1 & \text{if } S_{\text{fused}}(i, j) > \theta_{\text{sim}} \\
        0 & \text{otherwise}
    \end{cases}
    \label{eq:adjacency}
\end{equation}
The resulting matrix $\mathcal{S}$ represents the adjacency structure of the graph $\mathcal{G}$, where an edge signifies that two tokens are both semantically similar and spatially proximal.

\subsection{Pivot-based Diverse Token Selection}
\label{sec:pivot_selection}
To construct a token subset of size $n$ that is both representative and non-redundant, we introduce a novel selection algorithm named Pivot-based Diverse Token Selection. Its core mechanism is inspired by greedy approaches to the Maximal Independent Set (MIS) problem: it iteratively selects a high-value token and disqualifies its topological neighbors. This method leverages pre-computed importance scores $\mathcal{I}(h_i)$ to gauge the value of each token. Instead of merely selecting the top-scoring tokens, which could be highly redundant, our algorithm actively enforces diversity through a two-stage topological exclusion mechanism. First, based on a predefined pivot ratio $r_{pivot}$, it selects $n_p = \lfloor n \cdot r_{pivot} \rfloor$ high-importance "pivots" with the highest scores, followed by an iterative expansion to choose the remaining $n - n_p$ tokens while excluding topological neighbors to ensure structural diversity. With a relatively large default $r_{pivot}$ (0.7) and few retained tokens, this phase requires minimal iterations, adding less than 0.5\% to inference time. Detailed steps are provided in the appendix.

%% file: table/llava_v1.5.tex
\renewcommand{\multirowsetup}{\centering}
\definecolor{mygray}{gray}{.92}
\definecolor{ForestGreen}{RGB}{34,139,34}
\definecolor{Forestred}{RGB}{220,50,50}
\begin{table*}[!ht]
    \centering
    \vspace{-1mm}
    \setlength{\tabcolsep}{3.5pt}
    \renewcommand{\arraystretch}{0.9}
    \footnotesize
    \centering
    \scalebox{0.9}{
    \begin{tabular}{c | c c c c c c c c c | >{\centering\arraybackslash}p{1.0cm}}
        \toprule[1.5pt]
        \textbf{Method} & \textbf{GQA} & \textbf{MMB} & \textbf{MMB-CN} & \textbf{MME} & \textbf{POPE} & \textbf{SQA} & \textbf{VQA}$^{\text{V2}}$ & \textbf{VQA}$^{\text{Text}}$ & \textbf{VizWiz} &  \makecell[c]{\textbf{Avg}.}\\
        \hline
        \rowcolor{mygray}
        LLaVA-1.5-7B & \multicolumn{10}{c}{\textit{Upper Bound, 576 Tokens} \ $\textbf{(100\%)}$}\\
        \textcolor{gray}{Vanilla} & \textcolor{gray}{61.9} & \textcolor{gray}{64.7} & \textcolor{gray}{58.1} & \textcolor{gray}{1862} & \textcolor{gray}{85.9} & \textcolor{gray}{69.5} & \textcolor{gray}{78.5} & \textcolor{gray}{58.2} & \textcolor{gray}{50.0} &  \multirow{1}*{\textcolor{gray}{100\%}} \\
        \hline

        \rowcolor{mygray}
        LLaVA-1.5-7B & \multicolumn{10}{c}{\textit{Retain 192 Tokens} \ $\fg{(\downarrow 66.7\%)}$}\\
        ToMe \texttt{\scriptsize{(ICLR23)}} & 54.3 & 60.5 & - & 1563 & 72.4 & 65.2 & 68.0 & 52.1 & - & \multirow{1}*{88.5\%} \\
        FastV \texttt{\scriptsize{(ECCV24)}} & 52.7 & 61.2 & 57.0 & 1612 & 64.8 & 67.3 & 67.1 & 52.5 & 50.8 & \multirow{1}*{91.2\%} \\
        HiRED \texttt{\scriptsize{(AAAI25)}} & 58.7 & 62.8 & 54.7 & 1737 & 82.8 & 68.4 & 74.9 & 47.4 & 50.1 & 91.5\%     \\
        FitPrune \texttt{\scriptsize{(AAAI25)}} & 60.4 & 63.3 & 56.4 & 1831 & 83.4 & 67.8 & - & 57.4 & 50.9 & 98.2\% \\
        LLaVA-PruMerge \texttt{\scriptsize{(ICCV25)}} & 54.3 & 59.6 & 52.9 & 1632 & 71.3 & 67.9 & 70.6 & 54.3 & 50.1 & 90.8\% \\
        \multirow{1}*{SparseVLM \texttt{\scriptsize{(ICML25)}}} & 57.6 & 62.5 & 53.7 & 1721 & 83.6 & 69.1 & 75.6 & 56.1 & 50.5 & 96.3\% \\
        PDrop \texttt{\scriptsize{(CVPR25)}} & 57.1 & 63.2 & 56.8 & 1766 & 82.3 & 68.8 & 75.1 & 56.1 & 51.1 & 96.7\% \\
        VisionZip \texttt{\scriptsize{(CVPR25)}} & 59.3 & 63.0 & - & 1783 & 85.3 & 68.9 & 77.4 & 57.3 & - & 97.8\% \\

        DART \texttt{\scriptsize{(Arxiv25)}}& 60.0 & 63.6& 57.0 & 1856 & 82.8 & 69.8 & 76.7 & 57.4 & 51.2 & 98.8\% \\
        \rowcolor{cyan!7}
        \algname (Ours) & 60.8 & 64.3 & 57.1 & 1879 & 85.6 & 70.0 & 78.0 & 58.2 & 51.4 & \textbf{99.9\%}  \\
        \hline

        \rowcolor{mygray}
        LLaVA-1.5-7B & \multicolumn{10}{c}{\textit{Retain 128 Tokens} \ $\fg{(\downarrow 77.8\%)}$}\\
        ToMe \texttt{\scriptsize{(ICLR23)}} & 52.4 & 53.3 & - & 1343 & 62.8 & 59.6 & 63.0 & 49.1 & -  & \multirow{1}*{80.4\%} \\
        FastV \texttt{\scriptsize{(ECCV24)}} & 49.6 & 56.1 & 56.4 & 1490 & 59.6 & 60.2 & 61.8 & 50.6 & 51.3  & \multirow{1}*{86.4\%}\\
        HiRED \texttt{\scriptsize{(AAAI25)}} & 57.2 & 61.5 & 53.6 & 1710 & 79.8 & 68.1 & 73.4 & 46.1 & 51.3 & 90.2\% \\
        FitPrune \texttt{\scriptsize{(AAAI25)}} & 58.5 & 62.7 & 56.2 & 1776 & 77.9 & 68.0 & - & 55.7 & 51.7 & 96.4\% \\
        LLaVA-PruMerge \texttt{\scriptsize{(ICCV25)}} & 53.3 & 58.1 & 51.7 & 1554 & 67.2 & 67.1 & 68.8 & 54.3 & 50.3 & 88.8\%  \\
        SparseVLM
         \texttt{\scriptsize{(ICML25)}} & 56.0 & 60.0 & 51.1 & 1696 & 80.5 & 67.1 & 73.8 & 54.9 & 51.4 & 93.8\% \\
        PDrop \texttt{\scriptsize{(CVPR25)}} & 56.0 & 61.1 & 56.6 & 1644 & 82.3 & 68.3 & 72.9 & 55.1 & 51.0  & 95.1\% \\
        VisionZip \texttt{\scriptsize{(CVPR25)}} & 57.6 & 62.0 & - & 1763 & 83.2 & 68.9 & 75.6 & 56.8 & -  & 96.2\% \\
        DART \texttt{\scriptsize{(Arxiv25)}} & 58.7 & 63.2 & 57.5 & 1840 & 80.1 & 69.1 & 75.9 & 56.4 & 51.7 & 98.0\% \\
        \rowcolor{cyan!7}
        \algname (Ours) & 60.3 & 63.7 & 56.5 & 1850 & 85.1 & 69.3 & 77.4 & 58.0 & 51.8 & \textbf{99.2\% }\\
        \hline

        \rowcolor{mygray}
        LLaVA-1.5-7B & \multicolumn{10}{c}{\textit{Retain 64 Tokens} \ $\fg{(\downarrow 88.9\%)}$}\\
        ToMe \texttt{\scriptsize{(ICLR23)}} & 48.6 & 43.7 & - & 1138 & 52.5 & 50.0 & 57.1 & 45.3 & - & 70.1\%  \\
        FastV \texttt{\scriptsize{(ECCV24)}} & 46.1 & 48.0 & 52.7 & 1256 & 48.0 & 51.1 & 55.0 & 47.8 & 50.8  & 77.3\% \\
        HiRED \texttt{\scriptsize{(AAAI25)}} & 54.6 & 60.2 & 51.4 & 1599 & 73.6 & 68.2 & 69.7 & 44.2 & 50.2  & 87.0\% \\
        FitPrune \texttt{\scriptsize{(AAAI25)}} & 52.3 & 58.5 & 49.7 & 1556 & 60.9 & 68.0 & - & 51.2 & 51.1  & 87.8\% \\
        LLaVA-PruMerge \texttt{\scriptsize{(ICCV25)}} & 51.9 & 55.3 & 49.1 & 1549 & 65.3 & 68.1 & 67.4 & 54.0 & 50.1  & 87.4\% \\
        SparseVLM \texttt{\scriptsize{(ICML25)}} & 52.7 & 56.2 & 46.1 & 1505 & 75.1 & 62.2 & 68.2 & 51.8 & 50.1 & 84.6\% \\
        PDrop \texttt{\scriptsize{(CVPR25)}} & 41.9 & 33.3 & 50.5 & 1092 & 55.9 & 68.6 & 69.2 & 45.9 & 50.7  & 78.1\% \\
        VisionZip \texttt{\scriptsize{(CVPR25)}} & 55.1 & 60.1 & - & 1690 & 77.0 & 69.0 & 72.4 & 55.5 & -  & 92.7\% \\
        DART \texttt{\scriptsize{(Arxiv25)}}& 55.9 & 60.6 & 53.2 & 1765 & 73.9 & 69.8 & 72.4 & 54.4 & 51.6  & 93.7\% \\
        \rowcolor{cyan!7}
        \algname (Ours) & 57.9 & 61.9 & 55.6 & 1823 & 82.4 & 70.0 & 74.6 & 56.1 & 52.2  & \textbf{97.3\%} \\

        \hline
        \hline
        \rowcolor{mygray}
        LLaVA-Next-7B & \multicolumn{10}{c}{\textit{Upper Bound, 2880  Tokens} \ $\textbf{(100\%)}$}\\
         \textcolor{gray}{Vanilla} & \textcolor{gray}{64.2} & \textcolor{gray}{67.4} & \textcolor{gray}{60.6} & \textcolor{gray}{1851} & \textcolor{gray}{86.5} & \textcolor{gray}{70.1} & \textcolor{gray}{81.8} & \textcolor{gray}{64.9} & \textcolor{gray}{57.6} &    \textcolor{gray}{100\%} \\
          \hline
       \rowcolor{mygray}
        LLaVA-Next-7B & \multicolumn{10}{c}{\textit{Retain 320 Tokens} \ $\fg{(\downarrow 88.9\%)}$} \\

        FastV \texttt{\scriptsize{(ECCV24)}} & 55.9 & 61.6 & 51.9 & 1661 & 71.7 & 62.8 & 71.9 & 55.7 & 53.1  & 86.4\% \\
        
        HiRED \texttt{\scriptsize{(AAAI25)}} & 59.3 & 64.2 & 55.9 & 1690 & 83.3 & 66.7 & 75.7 & 58.8 & 54.2  & 91.8\% \\
        
        LLaVA-PruMerge \texttt{\scriptsize{(ICCV25)}} & 53.6 & 61.3 & 55.3 & 1534 & 60.8 & 66.4 & 69.7 & 50.6 & 54.0  & 79.9\% \\

       SparseVLM \texttt{\scriptsize{(ICML25)}} & 56.1 & 60.6 & 54.5 & 1533 & 82.4 & 66.1 & 71.5 & 58.4 & 52.0  & 85.9\%  \\

       PDrop \texttt{\scriptsize{(CVPR25)}} & 56.4 & 63.4 & 56.2 & 1663 & 77.6 & 67.5 & 73.5 & 54.4 & 54.1  & 86.8\% \\

       FasterVLM \texttt{\scriptsize{(Arxiv24)}} & 56.9 & 61.6 & 53.5 & 1701 & 83.6 & 66.5 & 74.0 & 56.5 & 52.6 & 89.8\% \\
       
       GlobalCom$^2$ \texttt{\scriptsize{(Arxiv25)}} & 57.1 & 61.8 & 53.4 & 1698 & 83.8 & 67.4 & 76.7 & 57.2 & 54.6& 90.3\%  \\
 
       DART \texttt{\scriptsize{(Arxiv25)}} & 61.7 & 64.2 & 58.2 & 1710 & 84.1 & 68.4 & 75.7 & 58.7 & 56.1 & 93.9\%  \\
       \rowcolor{cyan!7}
       \algname (Ours)& 62.3 & 64.9 & 57.2 & 1717 & 84.6 & 69.0 & 78.9 & 58.9 & 54.8  & \textbf{95.5\%}  \\

        \bottomrule[1.5pt]
	\end{tabular}}
        \vspace{-2mm}
	\caption{Performance comparisons on LLaVA-1.5-7B and LLaVA-Next-7B across several understanding benchmarks.}
    \label{tab:llavav1.5_understanding}
     \vspace{-10pt}
\end{table*}

%% file: table/ud.tex


\begin{table*}[htbp]
    \centering

    \resizebox{\textwidth}{!}{\setlength{\tabcolsep}{1.5pt}
    \renewcommand{\arraystretch}{1.25}
    \begin{tabular}{p{1.8 cm} | ccccccc|c || ccccccc|c}
        \toprule[1.5pt]
        
        \multirow{2}{*}{\textbf{Method}} & \multicolumn{8}{c||}{\textbf{Qwen2.5-VL-7B}} & \multicolumn{8}{c}{\textbf{InternVL-2.5-8B}} \\
        \cline{2-17}
        ~ & \textbf{AI2D} & \textbf{ChartQA} & \textbf{DocVQA} & \textbf{MME} & \textbf{MMStar}& \textbf{POPE} &\textbf{TextVQA}& \textbf{Avg.} & \textbf{MME} & \textbf{POPE} & \textbf{Nocaps} & \textbf{OKVQA} & \textbf{VizWiz}& \textbf{ Filckr30k} & \textbf{VQAv2} & \textbf{Avg.} \\
        \hline

        \rowcolor{mygray}
        ~ & \multicolumn{8}{c||}{\textit{Upper Bound, All Tokens} \ $\textbf{(100\%)}$} & \multicolumn{8}{c}{\textit{Upper Bound, All Tokens} \ $\textbf{(100\%)}$} \\
        \textcolor{gray}{Vanilla} 
        & \textcolor{gray}{74.1} & \textcolor{gray}{77.4} & \textcolor{gray}{91.0} & \textcolor{gray}{2240} & \textcolor{gray}{58.4} & \textcolor{gray}{87.7}  & \textcolor{gray}{76.23}& \textcolor{gray}{100\%}
        & \textcolor{gray}{2349} & \textcolor{gray}{90.49} & \textcolor{gray}{1.16} & \textcolor{gray}{64.56} & \textcolor{gray}{63.51} & \textcolor{gray}{95.99} & \textcolor{gray}{81.70} & \textcolor{gray}{100\%} \\
        \hline

        \rowcolor{mygray}
        ~ & \multicolumn{8}{c||}{\textit{Retain 25\% Tokens in Average} \ $(\downarrow 75\%)$} & \multicolumn{8}{c}{\textit{Retain 25\% Tokens in Average} \ $(\downarrow 75\%)$} \\
        FastV  
        & 67.49 & 63.68 & 71.26 & 2073 & 49.26 & 81.54 & 74.65 & 88.49\%
        & 2251 & 89.76 & 1.14 & 63.89 & 62.91 & 91.54 & 79.46 & 97.57\% \\
        DART
        & 62.37 & 49.44 & 52.99 & 2086 & 50.58 & 82.53 & 64.74 & 80.72\%
        & 2252 & 89.70 & 1.11 & 63.83 & 62.66 & 89.33 & 79.47 & 96.94\% \\ 
        DivPrune
        & 69.95 & 58.84 & 73.88 & 2097 & 52.56 & 83.98 & 70.43 & 89.05\%
        & 2189 & 89.91 & 1.11 & 63.46 & 62.29 & 89.29 & 79.23 & 96.37\% \\ 
        \algname  
        & 71.83 & 68.40 & 81.14 & 2190 & 54.50 & 85.22 & 74.64 & \textbf{94.38\%}
        & 2316 & 90.18 & 1.16 & 64.23 & 63.56 & 94.50 & 80.13 & \textbf{99.19\%} \\
        \hline

        \rowcolor{mygray}
        ~ & \multicolumn{8}{c||}{\textit{Retain 10\% Tokens in Average} \ $(\downarrow 90\%)$} & \multicolumn{8}{c}{\textit{Retain 10\% Tokens in Average} \ ($\downarrow 90\%$)} \\
        FastV  
        & 57.71 & 38.64 & 36.85 & 1692 & 37.4 & 65.52 & 66.40 & 67.10\%
        & 1999 & 85.05 & 1.02 & 59.99 & 59.65 & 75.76 & 69.65 & 88.31\% \\
        DART
        & 58.55 & 33.04 & 30.1 & 1853 & 42.08 & 71.29 & 48.31 & 64.89\%
        & 2057 & 81.53 & 0.90 & 59.94 & 60.33 & 66.64 & 69.91 & 85.45\% \\ 
        DivPrune
        & 62.73 & 39.28 & 49.34 & 1909 & 45.10 & 77.97 & 58.79 & 74.01\%
        & 2081 & 88.29 & 1.00 & 61.63 & 60.56 & 78.50 & 74.96 & 90.96\% \\ 
        \algname  
        & 64.73 & 51.96 & 57.54 & 2004 & 48.33 & 78.34 & 70.54 & \textbf{81.69\%}
        & 2196 & 88.23 & 1.17 & 62.48 & 63.61 & 90.05 & 76.92 & \textbf{96.67\%} \\
        
        \bottomrule[1.5pt]
    \end{tabular}}
    
    \caption{Performance comparisons on Qwen2.5-VL-7B and InternVL-2.5-8B across several understanding benchmarks}
    \label{tab:ud}

\end{table*}

%% file: table/vg.tex
\renewcommand{\multirowsetup}{\centering}
\begin{table*}[t]
\renewcommand{\arraystretch}{1.3}
\vspace{2pt}
\centering
\setlength{\tabcolsep}{4pt} 
\resizebox{\textwidth}{!}{
\begin{tabular}{l|ccc|ccc|cc|c||ccc|ccc|cc|c}
\noalign{\hrule height 1pt}

\multirow{3}{*}{\textbf{Method}} & \multicolumn{9}{c||}{\textbf{LLaVA-1.5-7B}} & \multicolumn{9}{c}{\textbf{InternVL-2.5-8B}} \\
\cline{2-19}
~ & \multicolumn{3}{c|}{\textbf{RefCOCO}} & \multicolumn{3}{c|}{\textbf{RefCOCO+}} & \multicolumn{2}{c|}{\textbf{RefCOCOg}} & \multirow{2}{*}{\textbf{Average}} & \multicolumn{3}{c|}{\textbf{RefCOCO}} & \multicolumn{3}{c|}{\textbf{RefCOCO+}} & \multicolumn{2}{c|}{\textbf{RefCOCOg}} & \multirow{2}{*}{\textbf{Average}} \\
\cline{2-9} \cline{11-18}
~ & \texttt{val} & \texttt{testA} & \texttt{testB} & \texttt{val} & \texttt{testA} & \texttt{testB} & \texttt{val} & \texttt{test} & ~ & \texttt{val} & \texttt{testA} & \texttt{testB} & \texttt{val} & \texttt{testA} & \texttt{testB} & \texttt{val} & \texttt{test} & ~ \\
\noalign{\hrule height 1pt}


\rowcolor{gray!20}
~ & \multicolumn{9}{c||}{\textit{Upper Bound, All Tokens (100\%)}} & \multicolumn{9}{c}{\textit{Upper Bound, All Tokens (100\%)}} \\
Upper Bound & 
\textcolor{gray!80}{69.29} & \textcolor{gray!80}{77.07} & \textcolor{gray!80}{59.88} & \textcolor{gray!80}{57.37} & \textcolor{gray!80}{69.56} & \textcolor{gray!80}{45.53} & \textcolor{gray!80}{65.36} & \textcolor{gray!80}{65.43} & \textcolor{gray!80}{100.0\%} &
\textcolor{gray!80}{90.16} & \textcolor{gray!80}{94.54} & \textcolor{gray!80}{85.98} & \textcolor{gray!80}{85.05} & \textcolor{gray!80}{91.56} & \textcolor{gray!80}{78.77} & \textcolor{gray!80}{87.03} & \textcolor{gray!80}{87.72} & \textcolor{gray!80}{100.0\%} \\
\noalign{\hrule height 1pt}

\rowcolor{gray!20}
~ & \multicolumn{9}{c||}{\textit{Retain 75\% Tokens in Average ($\downarrow$ 25\%)}} & \multicolumn{9}{c}{\textit{Retain 50\% Tokens in Average ($\downarrow$ 50\%)}} \\
FastV & 65.94 & 73.47 & 57.61 & 54.93 & 65.86 & 43.1 & 61.83 & 61.84 & 95.11\% & 88.07 & 92.15 & 83.77 & 82.68 & 88.91 & 75.95 & 85.01 & 85.63 & 97.33\% \\
DivPrune & 43.78	&48.56	&38.98	&35.23	&41.83	&27.45	&39.89	&39.02 & 61.74\% & 87.06	&91.32	&81.77	&81.44	&88.47	&74.37	&83.95	&84.42& 95.97\% \\
DART & 65.54 & 73.01 & 57.74 & 54.17 & 64.81 & 44.00 & 62.38 & 61.84 & 95.11\% & 86.15 & 91.32 & 81.73 & 80.91 & 87.72 & 74.13 & 83.37 & 84.1 & 95.49\% \\
\algname & 68.37 & 76.15 & 59.33 & 56.91 & 68.06 & 44.22 & 64.07 & 63.03 & \textbf{98.14\%} & 88.97 & 93.44 & 84.73 & 83.42 & 90.57 & 77.71 & 85.85 & 86.03 & 98.43\%\\
\noalign{\hrule height 1pt}

\rowcolor{gray!20}
~ & \multicolumn{9}{c||}{\textit{Retain 50\% Tokens in Average ($\downarrow$ 50\%)}} & \multicolumn{9}{c}{\textit{Retain 25\% Tokens in Average ($\downarrow$ 75\%)}} \\
FastV & 56.36 & 63.27 & 48.32 & 46.13 & 55.34 & 35.96 & 52.96 & 51.6 & 80.39\% & 78.17 & 83.38 & 72.91 & 71.68 & 77.98 & 65.04 & 74.96 & 76.37 & 85.64\% \\
DivPrune &20.11	&24.02	&18.41	&16.1	&20.47	&12.78	&18.22	&18.02& 28.99\% & 73.28	&78.56	&68.44	&67.75	&75.22	&61.44	&70.73	&71.5 & 80.96\% \\
DART & 53.54 & 60.62 & 47.22 & 44.79 & 53.89 & 36.69 & 51.41 & 51.17 & 78.49\% & 66.83 & 70.94 & 62.65 & 61.54 & 67.69 & 56.99 & 66.83 & 67.56 & 74.33\% \\
\algname & 62.7 & 70.64 & 54.29 & 51.53 & 62.57 & 40.7 & 58.44 & 57.81 & \textbf{90.00\%} & 86.09 & 89.78 & 82.08 & 79.8 & 85.85 & 74.56 & 82.23 & 83.27 & 94.71\%\\
\noalign{\hrule height 1pt}

\rowcolor{gray!20}
~ & \multicolumn{9}{c||}{\textit{Retain 40\% Tokens in Average ($\downarrow$ 60\%)}} & \multicolumn{9}{c}{\textit{Retain 10\% Tokens in Average ($\downarrow$ 90\%)}} \\
FastV & 49.10 & 56.19 & 42.73 & 40.47 & 48.13 & 31.72 & 46.26 & 44.99 & 70.76\% & 31.98 & 36.63 & 28.13 & 27.1 & 31.52 & 23.6 & 32.27 & 31.26 & 34.50\% \\
DivPrune & 12.80	&14.77	&12.68	&10.19	&12.00	&9.16	&12.32	&11.60 & 18.82\% & 36.99	&39.12	&35.72	&32.87	&35.61	&30.84	&35.19	&36.94 & 40.65\% \\
DART & 45.69 & 52.87 & 39.02 & 37.67 & 45.95 & 30.31 & 44.36 & 43.80 & 66.87\% & 19.51 & 20.35 & 20.22 & 16.78 & 17.99 & 17.37 & 20.94 & 21.93 & 22.15\% \\
\algname & 59.59 & 68.07 & 52.22 & 48.56 & 59.87 & 38.23 & 57.03 & 54.63 & \textbf{85.89\%} & 78.21 & 82.69 & 73.99 & 72.05 & 78.03 & 65.56 & 75.12 & 75.02 & 85.68\% \\
\noalign{\hrule height 1pt}

\end{tabular}

}
\vspace{-5pt}
\caption{Performance comparisons on LLaVA-1.5-7B and InternVL2.5-8B across localization benchmarks. }
\vspace{-10pt}
\label{tab:vg}
\end{table*}

%% file: section/4_experiment.tex
\section{Experiments}

\input{image/ablation}
\subsection{Experimental Settings}
\paragraph{Models and Baselines.}
We apply D²Prune to four popular MLLMs with different architectures to
evaluate its general effectiveness. Specifically, we follow previous work in this field to compare performance on LLaVA-1.5-7B~\cite{llava1.5} and LLaVA-NeXT-7B~\cite{llavanext}. Furthermore, we also present experimental results on the recent Qwen-2.5-VL-7B~\cite{qwenvl2.5} and InternVL-2.5-8B~\cite{internvl2.5}. We compare the performance of our approach with multiple token reduction methods: ToMe~\cite{tome}, FastV~\cite{fastv}, HiRED~\cite{hired}, FitPrune~\cite{fitprune}, LLaVA-PruMerge~\cite{llava-prumerge}, SparseVLM~\cite{sparsevlm}, HiRED~\cite{hired}, PyramidDrop~\cite{pdrop}, VisionZip~\cite{visionzip}, FasterVLM~\cite{fastervlm}, GlobalCom$^3$~\cite{globalcom} DivPrune~\cite{divprune} and DART~\cite{dart}.

\paragraph{Benchmarks.}
We perform comprehensive evaluations across general understanding benchmarks, which include visual understanding and reasoning datasets such as GQA~\cite{gqa}, ScienceQA~\cite{sqa}, VQAv2~\cite{vqav2}, TextVQA~\cite{vqatext}, and VizWiz~\cite{vizwiz}, as well as multi-modal reasoning benchmarks like MMBench~\cite{mmbench}, MMBench-CN~\cite{mmbench}, MME~\cite{mme}, POPE~\cite{pope}, AI2D~\cite{ai2d}, ChartQA~\cite{chartqa}, DocVQA~\cite{docvqa}, OKVQA~\cite{okvqa}, MMStar~\cite{mmstar}, Filckr30k~\cite{flickr30k} and Nocaps~\cite{nocaps}. Our experiments also extend to more challenging referring grounding tasks, utilizing RefCOCO\cite{refcoco}, RefCOCO+~\cite{refcoco}, and RefCOCOg~\cite{refcocog}.

\paragraph{Implementation Details.}
We follow the default inference settings provided in the official codebase for each MLLM. For our method's hyperparameters, pruning is initiated at layer $K=2$ for all scenarios except for the LLaVA-based localization task, where $K$ is set to 5. The weighting parameter $\alpha$ is assigned a value of 1 for understanding tasks and 0.5 for localization tasks. By default, 
$r_{pivot}$ is set to 0.7 and $\theta_{sim}$ is set to 0.8.

\subsection{Main Results}
\paragraph{Results on understanding task.} We present the performance on understanding tasks in Tab.~\ref{tab:llavav1.5_understanding} and Tab.~\ref{tab:ud}.
For \textbf{LLaVa-1.5-7B}, D²Pruner achieves an impressive average performance of 99.9\%, with only 192 tokens retained (↓ 66.7\%). 
When further reducing token retention to 64 tokens (↓ 77.8\%), D²Pruner still maintains a strong performance of 97.3\%, demonstrating its robustness even with a sharp reduction in tokens. 
For \textbf{LLaVa-Next-7B}, the D²Pruner achieves exceptional scalability, achieving an average score of 95.5\% with only 11.1\% of tokens retained, surpassing all other methods by a notable margin.
For \textbf{Qwen2.5-VL-7B}, with 90\% token reduction, D²Pruner achieves an impressive 81.69\%, outperforming DART (64.89\%), FastV (67.10\%) and DivPrune (74.01\%) by notable margins. 
For \textbf{InternVL2.5-8B}, D²Pruner prunes a substantial 75\% of tokens while retaining 99.19\% of the original performance, demonstrating a nearly lossless compression. When reducing tokens by 90\%, D²Pruner still delivers 96.67\%, outpacing all other methods by a wide margin.
\paragraph{Results on localization task.}
We futher evaluate our method and compare it with several previous methods on three widely-used referring expression comprehension benchmarks: RefCOCO, RefCOCO+, and RefCOCOg. The results are summarized in Tab.~\ref{tab:vg}.
For \textbf{LLaVA-1.5-7B}, under a high 60\% reduction ratio, our method's average performance of 85.89\% surpasses that of other methods by 15.13-67.07 percentage points. The robustness of our method is particularly evident on \textbf{InternVL2.5-8B}. Even when 90\% of tokens are pruned, our method maintains a high performance of 85.68\%, in stark contrast to competing methods, whose performance drops to a mere 22.1\%-40.65\%.

\subsection{Ablation Studies}
\paragraph{Effectiveness of DI and SD.} We conduct experiments on the understanding task with Qwen2.5-VL-7B~\cite{qwenvl2.5} and the localization task with LLaVA-1.5-7B~\cite{llava1.5}, comparing our full method (\textit{Ours (DI+SD)}) against the FastV~\cite{fastv} baseline and a variant using only Debiased Importance (\textit{Ours (DI)}). As shown in Fig.~\ref{fig:ablation} (left), Ours (DI) already surpasses the baseline, demonstrating that correcting for attention bias is effective. The addition of Structural Diversity (\textit{Ours (DI+SD)}) further improves performance by encouraging diversity among the selected tokens, which leads to significant gains on both tasks. This confirms that DI successfully identifies critical tokens, while SD complements it by promoting diversity in token selection. Notably, the performance advantage of D²Pruner becomes more pronounced at higher pruning ratios, highlighting the robustness and synergistic effect of our two components under extreme compression.

\paragraph{Effectiveness of spatial adjacency.}  
To evaluate the impact of spatial adjacency, we compare our method with and without this component on LLaVA-1.5-7B under a reduction ratio of 0.5. As shown in Fig.~\ref{fig:ablation}(right), incorporating spatial adjacency consistently improves performance on localization tasks. This demonstrates that considering spatial redundancy and ensuring a more uniform spatial distribution are crucial for localization, as they help preserve important spatial cues after token reduction.

\subsection{Efficiency Analysis}
Our method provides substantial and flexible efficiency gains, as shown in Tab.\ref{tab:efficiency} for LLaVA-Next-7B. Retaining just 33.4\% of tokens accelerates prefilling by 2.93× and cuts FLOPs/KV Cache by over 2.8×. Under a more aggressive 11.2\% keep ratio of visual tokens, we achieve a 5.09× prefill speedup and reduce FLOPs/KV Cache by 6.10×/7.63×. Notably, the entire pruning process is highly efficient, incurring a negligible overhead of just \textbf{0.8\%} of a single forward pass.
\input{table/efficiency}

%% file: image/ablation.tex
\begin{figure*}[t]
\centering
\small
\includegraphics[width=0.99\textwidth]{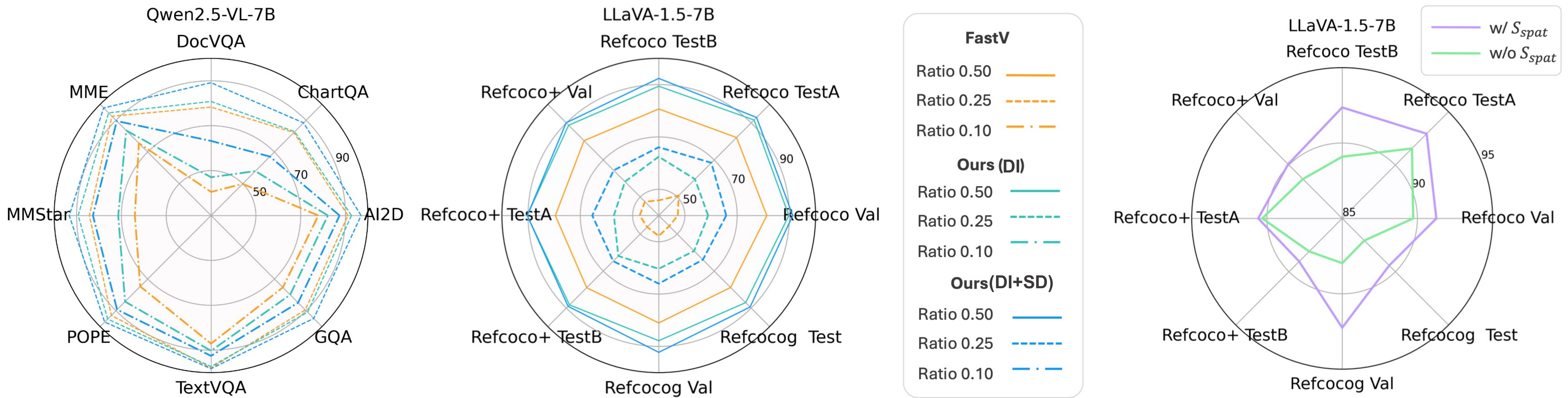}
\vspace{-5 pt}
\caption{\textbf{Ablation study.}  \textit{Left}: Effectiveness of Debiased Importance (DI) and Structural Diversity (SD). Our full method (DI+SD) consistently outperforms FastV and the DI-only variant, especially at low token rataining ratios (e.g., 0.25, 0.1). \textit{Right}: Effectiveness of spatial adjacency. Our method with $S_{spat}$ outperforms the variant without it, demonstrating the benefit of considering spatial distribution for localization tasks.}
\vspace{-10pt}
\label{fig:ablation}
\end{figure*}

%% file: table/efficiency.tex
\begin{table}[!ht]
    \centering
    \setlength{\tabcolsep}{2.0pt}
    \renewcommand{\arraystretch}{0.95}
    \footnotesize
    \scalebox{0.7}{
    \begin{tabular}{@{}lccccc}
        \toprule[1.2pt]
         \multirow{2}{*}{\textbf{Methods}}  & \textbf{Prefilling Time $\downarrow$} & \textbf{Total Time $\downarrow$} & \textbf{FLOPs $\downarrow$} & \textbf{KV Cache $\downarrow$}  & \multirow{1}{*}{\textbf{POPE $\uparrow$}}   \\
           & \textbf{(ms/sample)} & \textbf{(ms/sample)} & \textbf{(T)}& \textbf{(MB)} & \textbf{(F1-Score)} \\
         \midrule
         \textcolor{gray}{LLaVA-Next-7B} & \textcolor{gray}{350 (1.00$\times$)} & \textcolor{gray}{411 (1.00$\times$)} & \textcolor{gray}{16.9 (1.00$\times$)} & \textcolor{gray}{1512 (1.00$\times$)} &\textcolor{gray}{86.5} \\
         +Ours(33.4\%)  & 119 (2.93$\times$) & 183 (2.24$\times$) & 6.0 (2.82$\times$)& 526 (2.87$\times$) & 85.4 \\
         +Ours(11.2\%)  & 68 (5.09$\times$) & 132 (3.10$\times$) & 2.8 (6.10$\times$)& 198 (7.63$\times$) & 84.6 \\
        
        \bottomrule[1.2pt]
    \end{tabular}}
    \vspace{-5pt}
    \caption{\textbf{Efficiency comparisons on LLaVA-Next-7B.}
We report the actual prefillling time, total runtime, theoretical FLOPs, KV cache, and F1-Score on the POPE benchmark.}

    \label{tab:efficiency}
    \vspace{-10pt}
\end{table}

%% file: section/5_conclusion.tex
\section{Conclusion}
In this paper, we first conduct a comprehensive analysis of existing token pruning strategies. We reveal that importance-based methods suffer from positional bias, while diversity-based methods exhibit structural blindness. To address these issues, we propose D²Pruner, a framework combining Debiased Importance (DI) and Structural Diversity (SD) to retain both critical and diverse tokens. Experimentally, D²Pruner achieves near-original performance (99.2\%) on general tasks with a 74.2\% FLOPs reduction. Critically, it surpasses prior methods by up to 63.5\% on challenging localization tasks, maintaining 85.7\% accuracy even when pruning 90\% of visual tokens from InternVL-2.5-8B.